\documentclass{bmvc2k}

\usepackage{amsmath}
\usepackage{amssymb}
\usepackage{multirow}
\usepackage{bbding}
\usepackage{pifont}
\usepackage{caption}
\usepackage{xcolor}

\title{Improving Object Detection from Scratch via Gated Feature Reuse}

\addauthor{Zhiqiang Shen$^1$\\ Honghui Shi$^{2,3}$ \\Jiahui Yu$^2$\\ Hai Phan$^1$\\ Rogerio Feris$^{3,4}$\\ Liangliang Cao$^5$ \\Ding Liu$^2$\\ Xinchao Wang$^6$\\ Thomas Huang$^2$\\ Marios Savvides}{{zhiqians,haithanp,marioss}@andrew.cmu.edu;{hshi10,jyu79,dingliu2,t-huang1}@illinois.edu}{1}
\addinstitution{
 Carnegie Mellon University\\
}
\addinstitution{
University of Illinois at Urbana-Champaign\\
}
\addinstitution{
 IBM Research AI\\
}
\addinstitution{
	MIT-IBM Watson AI Lab\\
}
\addinstitution{
Google AI \& UMass Amherst\\
}
\addinstitution{
Stevens Institute of Technology\\
}
\runninghead{Shen et al.}{Improving Object Detection via Gated Feature Reuse}

\begin{document}

\maketitle

\begin{abstract}
In this paper, we present a simple and parameter-efficient drop-in module for one-stage object detectors like SSD~\cite{liu2016ssd} when learning from scratch (i.e., without pre-trained models). We call our module GFR (\textbf{G}ated \textbf{F}eature \textbf{R}euse), which 
exhibits two main advantages. First, we introduce a novel gate-controlled prediction strategy enabled by Squeeze-and-Excitation~\cite{hu2017squeeze} to adaptively enhance or attenuate supervision at different scales based on the input object size. As a result, our model is more effective in detecting diverse sizes of objects. Second, we propose a feature-pyramids structure to squeeze rich spatial and semantic features into a single prediction layer, which strengthens feature representation and reduces the number of parameters to learn. We apply the proposed structure on DSOD~\cite{Shen2017DSOD,8734700} and SSD~\cite{liu2016ssd} detection frameworks, and evaluate the performance on PASCAL VOC 2007, 2012, 2012 \texttt{Comp3} and COCO datasets. With fewer model parameters, GFR-DSOD outperforms the  baseline DSOD by 1.4\%, 1.1\%, 1.7\% and 0.6\%, respectively. GFR-SSD also outperforms the original SSD and SSD with dense prediction by 3.6\% and 2.8\% on VOC 2007 dataset. Code: \url{https://github.com/szq0214/GFR-DSOD}. 
\end{abstract}

\section{Introduction}
\label{sec:intro}
\begin{figure}[t]
	\centering
    ~~~~~~~~~~~~~~~~~~
	\includegraphics[width=0.74\textwidth]{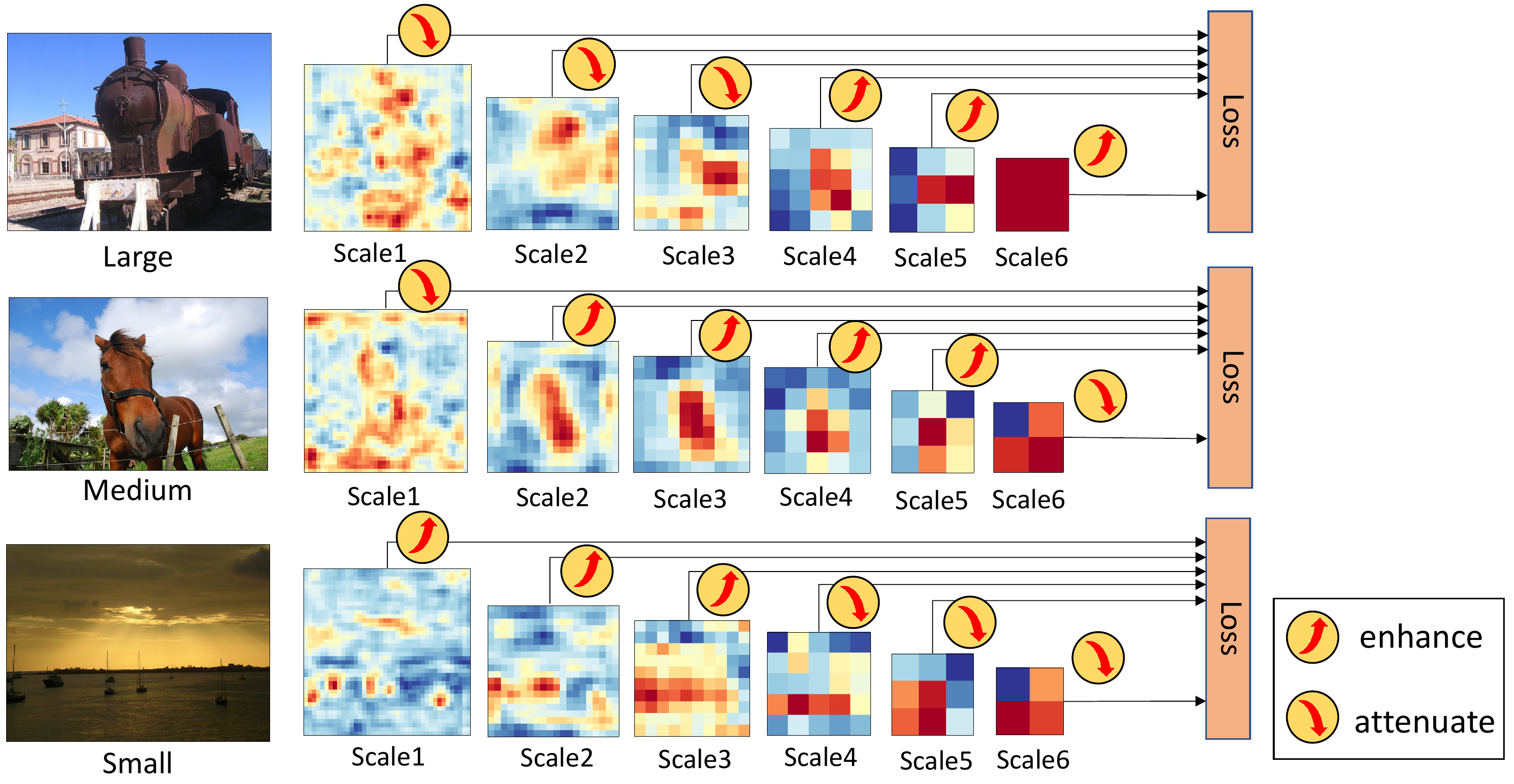}
	\vspace{0.08in}
	\caption{Illustration of the motivation. For a wide range of input object sizes, we adopt gates (the arrows upon the feature maps) on each scale to adaptively recalibrate the supervision intensities based on the input object sizes. Specifically, for large or small objects, our gates will automatically enhance or attenuate feature activations at appropriate scales.}
	\label{mot}
	\vspace{-0.1in}
\end{figure}

In the recent years, with the wide usage of many electronic devices such as mobile phones, embedded devices, etc, learning compact object detectors with high accuracy has been a urgent-needed breakthrough problem in computer vision 
and a promising research direction in deep learning.
However, most existing state-of-the-art object detection systems are designed based on backbone networks that are pre-trained on classification datasets (e.g., ImageNet~\cite{krizhevsky2012imagenet}), which are  fairly heavy on \#params and inflexible to adjust appropriate structures for different scenarios and circumstances. In this paper, we tackle this problem by considerring the following two aspects: (i) how to improve utilization of features on different scales; and (ii) how to detect diverse sizes of objects with gate-controlled prediction strategy, in order to build a parameter-efficient object detector. Our design in this paper can further bring  incidental advantage that our model is easier to train from scratch as we use fewer parameters and more simple structures in our models.

Some recent works have been explored to learn lite and parameter-efficient network architectures for image classification task, such as MoblileNet~\cite{howard2017mobilenets}, MoblileNet V2~\cite{sandler2018mobilenetv2}, ShuffleNet~\cite{zhang2018shufflenet}, etc. In our work, we adopt the existing backbone networks and focus on designing the detection head, our gated structure can be used in conjunction with any other backbone structure techniques.
There are also some recent studies focusing on developing feature-pyramids structures for detectors which have 
achieved very promising results. One example is the FPN~\cite{lin2016feature} model.
Similar to many other  detection frameworks such as SSD~\cite{liu2016ssd}, DSSD~\cite{fu2017dssd}, DSOD~\cite{Shen2017DSOD,8734700}, Reconfig~\cite{kong2018deep}, MS-CNN~\cite{cai2016unified}, Hypernet~\cite{kong2016hypernet} and ION~\cite{bell2016inside},  
FPN relies on \textit{feature pyramids} for multiple scale prediction.
For different pyramidal layers, a series of prediction operations are conducted to adapt the arbitrary object scales.

The current {\em feature pyramids} design, however, has two major limitations. 
First is that 
each pyramid has a fixed and thus non-adaptive contribution to the final supervision signals. 
Intuitively, objects at small scales may be easier to detect with fine-resolution (lower-level) features, 
and thus signals from those lower-level features should be enhanced; 
Similarly, large scale objects could be easier to detect at coarse-resolution (higher-level) feature maps, 
and thus signals from those higher-level feature maps should be enhanced.
Nevertheless, these are unfortunately not taken into account by the state-of-the-art detector pyramids.
The second limitation is the naive single-scaled feature representation in each pyramidal layer, as done in SSD~\cite{liu2016ssd} and FPN~\cite{lin2016feature}, where the pyramidal layers are independent without any interactions.

In this paper, we introduce a novel gated feature reuse (GFR) mechanism
that explicitly tackles the above two challenges.
To address the first challenge, we design a gate structure enabled by Squeeze-and-Excitation~\cite{hu2017squeeze} that dynamically adjusts  supervision at different scales. 
As illustrated in Fig.~\ref{mot},
for large or small objects, our gates automatically intensify or diminish
the feature activations at proper scales.
For the second challenge, we propose {\em feature reuse}, 
a network that concatenates high-level semantic features and low-level 
spatial features in a single pyramid,
so that the features may be complementary with each other and jointly lead to better results.
Our ablation experiments show that this simple design vastly 
boost the performance of object detection  and reduce the model parameters.
We incorporate our proposed GFR mechanism into DSOD and SSD, 
resulting in higher detection accuracy, fewer parameters, and faster convergence, as shown in Fig. {\color{red}{6}}. Such incorporation is done without bells and whistles, 
and could be readily extended to other detection frameworks.

Our main contributions are summarized as follows:
\vspace{-1.5mm}
\begin{itemize}
	\addtolength{\itemsep}{-0.11in}
	\item[(1)] We propose the {\em Feature Reuse}, a novel structure for learning detectors from scratch,
	that allows the features of different scales to interact, further leading to less model parameters and faster convergence speed.
	\item[(2)] We propose {\em Gating Mechanism} for generic object detection, which is, to our best knowledge,
	 the first successful attempt on recalibrating supervision signals for detection.
	\item[(3)] We apply {\em GFR} structure on two detection frameworks DSOD and SSD, the resulting GFR-DSOD method achieves state-of-the-art (under similar \#parames) for learning detection from scratch with real-time processing speed and more compact models.
\end{itemize}

\vspace{-3 ex}
\section{Related Work}

\noindent{\textbf{Object Detection.}} 
Generally, modern object detection frameworks fall into two groups. One is the two-stage detectors like R-CNN~\cite{girshick2014rich}, Fast RCNN~\cite{girshick2015fast}, R-FCN~\cite{dai2016r}, Faster RCNN~\cite{ren2015faster}, Deformable CNN~\cite{dai2017deformable}, Mask RCNN~\cite{he2017mask}, etc. In this paper we focus on another group: one-stage detectors (also known as proposal-free detectors). OverFeat~\cite{sermanet2013overfeat} is regarded as one of the first CNN-based one-stage detectors. After that, a number of recent detectors have been proposed, such as YOLO~\cite{redmon2016you,redmon2016yolo9000}, SSD~\cite{liu2016ssd}, RetinaNet~\cite{lin2017focal}, DSOD~\cite{Shen2017DSOD,8734700}, CornerNet~\cite{law2018cornernet}, ExtremeNet~\cite{zhou2019bottom}, Refinement~\cite{zhang2018single}, UnitBox~\cite{yu2016unitbox}, STDN~\cite{zhou2018scale}, RFB~\cite{liu2018receptive}, etc. 
The key advantages of one-stage detectors is the straight-forward structures and high speed but their accuracy is moderate. Thus, the aim of this work is to further boost the performance of one-stage detectors under the setting of training from scratch.

\begin{figure*}[t]
	\centering
	\includegraphics[width=0.78\textwidth]{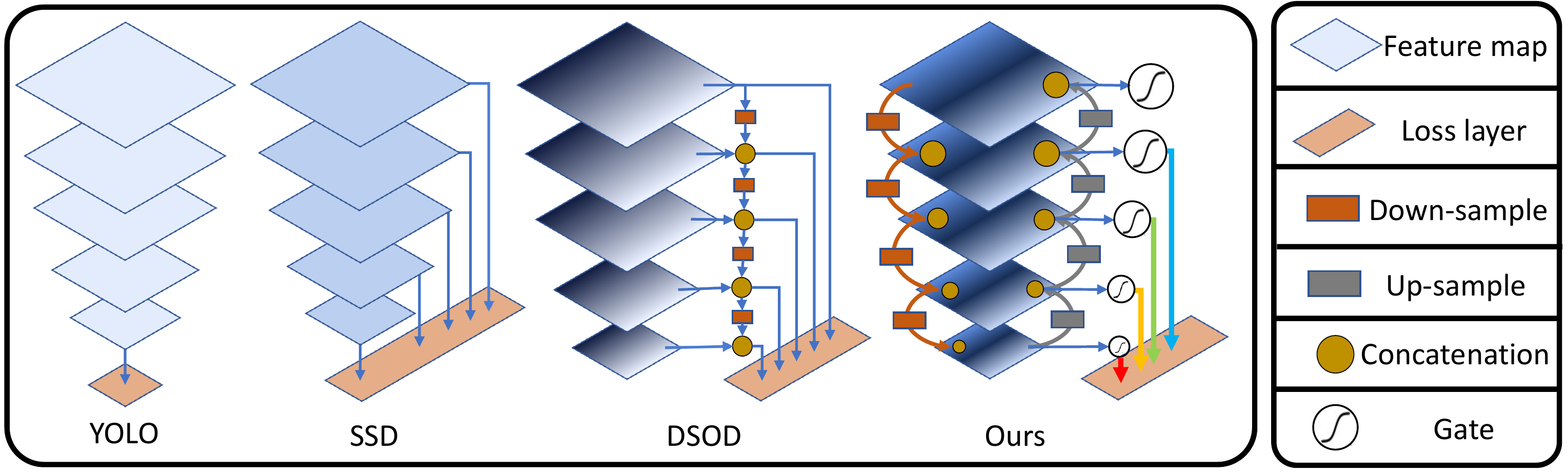}
	\vspace{0.1in}
	\caption {An overview of our proposed GFR-DSOD together with three one-stage detector methods (YOLO~\cite{redmon2016you}, SSD~\cite{liu2016ssd} and DSOD~\cite{Shen2017DSOD,8734700}). YOLO is a single-scale detector, SSD and DSOD are two multi-scale detectors that localize objects based on various resolution features. DSOD further adopts dense connections on prediction layers to combine different resolution features into one scale. Our GFR-DSOD consists of two modules: {\em Iterative Feature Pyramids} and {\em Gating Mechanism}. More details are given in \S\ref{sec:method}.}
	\label{Paradigms}
	\vspace{-0.1in}
\end{figure*}

\noindent{\textbf{Feature Pyramids.}}
Adopting multiple layers for generic object detection is a common practice in a range of recently proposed approaches~\cite{Shen2017DSOD,lin2016feature,liu2016ssd,fu2017dssd}. For instance, DSOD~\cite{Shen2017DSOD,8734700} applies dense connections in prediction layers to combine different resolution features for detecting objects. FPN~\cite{lin2016feature} develops a top-down architecture with lateral connections to build pyramidal hierarchy semantic features at all scales. DSSD~\cite{fu2017dssd} involves extra deconvolutional layers to capture additional large-scale context. In this paper, we propose a novel iterative feature pyramids structure to not only improve accuracy but also reduce parameters.

\noindent{\textbf{Gating Mechanism.}}
Gating (or attention) can be viewed as a process to adaptively adjust or allocate resource intensity towards the most informative or useful components of inputs. There are several methods for exploiting gating mechanism to improve image classification~\cite{srivastava2015highway,wang2017residual,hu2017squeeze} and detection~\cite{zeng2016gated}. GBD-Net~\cite{zeng2016gated} proposes a gated bi-directional CNN for object detection that passes messages between features from different regions and uses gated functions to control message transmission. SENet~\cite{hu2017squeeze} uses gating mechanism to model channel-wise relationships and enhances the representation power of modules throughout the networks. In this paper, we introduce an effective gating mechanism that combines attention both locally and globally.

\vspace{-1 ex}
\section{Method} \label{sec:method}

We begin with presenting Iterative Feature Reuse, a structure that combines adjacent layer features iteratively for object detection. 
Following that, we introduce how an elaborately designed gating mechanism is used to adaptively control supervision at multiple scales in a deep network.
Finally, we show how the above two structures can be applied in DSOD~\cite{Shen2017DSOD,8734700} and SSD~\cite{liu2016ssd} seamlessly to obtain GFP-DSOD~\cite{Shen2017DSOD}/SSD~\cite{liu2016ssd}.

\begin{figure}[t]
	\centering
	\includegraphics[width=0.5\textwidth]{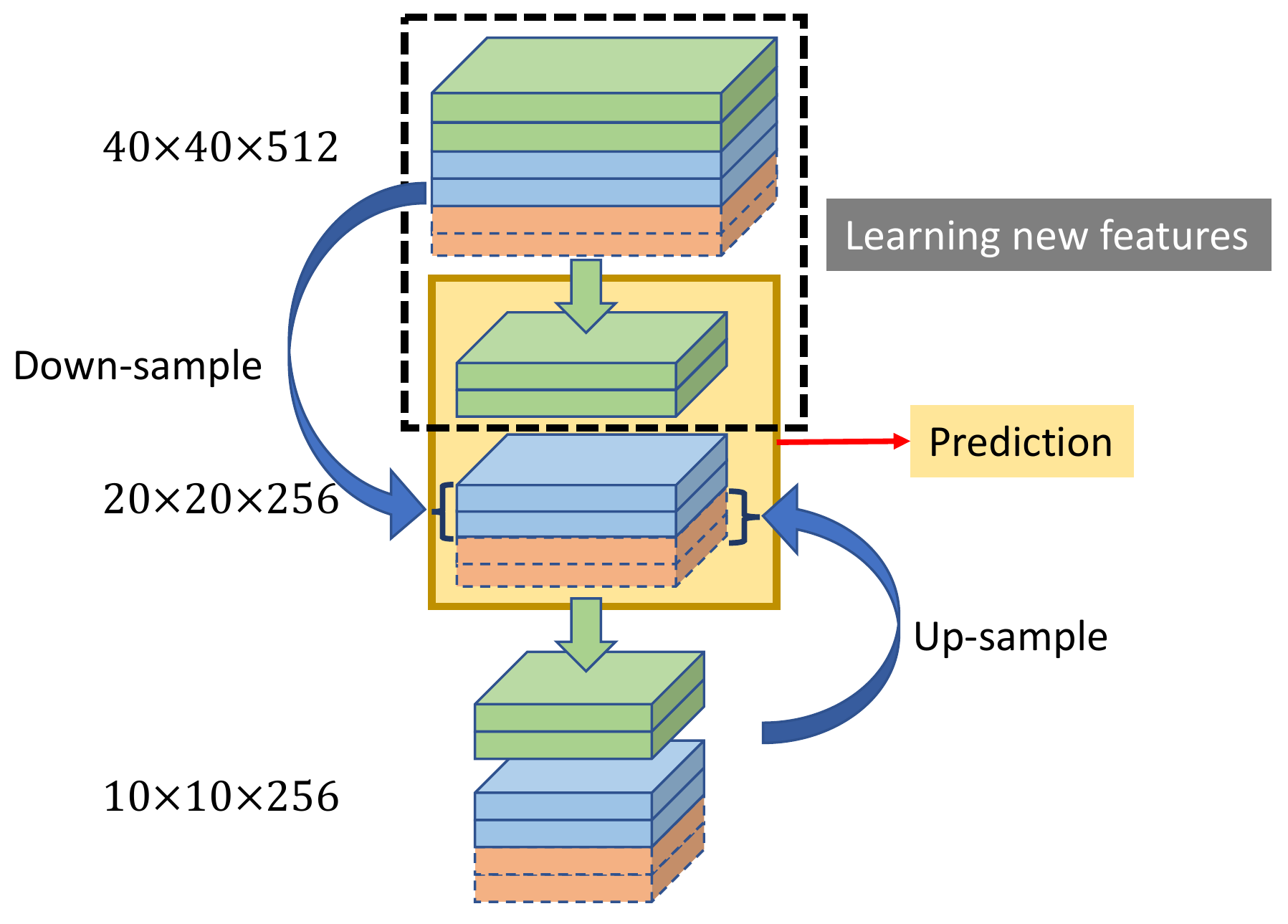}
	\vspace{0.1in}
	\caption{A building block illustrating the iterative feature pyramids, including down-sampling pathway, up-sampling pathway and the concatenation operation. Green color maps are the learned new features, blue and red ones are the re-used old features from previous and following layers, respectively.}
	\label{recurrent}
	\vspace{-0.1in}
\end{figure}

\vspace{-1 ex}
\subsection{Iterative Feature Re-Utilization}
Our goal is to utilize multi-scale CNN features in a single prediction pass that combines detailed shape and appearance cues from lower level layers and semantics from higher levels. 
Our proposed feature reuse structure is split into three parts, shown in Fig.~\ref{recurrent}. First, a down-sampling pathway takes the low level feature maps, and through a down-sampling block outputs the low-resolution feature maps that should be concatenated with the current features. Then, to incorporate higher-level semantics, an up-sampling pathway is adopted to concatenate high level features to the current layer. Finally, we repeat the concatenation operation in each scale of prediction layers in an iteration fashion.

\noindent{\textbf{Down/Up-Sampling Pathways.}}
The down-sampling pathway consists mainly of a max-pooling layer ($kernel\ size=2\times2$, $stride=2$), followed by a conv-layer ($kernel\ size=1\times1$, $stride=1$) to reduce channel dimensions, which is similar to the DSOD down-sampling block. The up-sampling pathway generates higher resolution features by upsampling spatially coarser, but semantically stronger features from the adjacent scale (we use nearest resolution in the upper layer for simplicity). We conduct a deconvolutional operation via bilinear upsampling followed by a conv-layer ($kernel\ size=1\times1$, $stride=1$) on the spatial resolution features maps. The upsampled maps are then concatenated with features from the down-sampling pathway and the current layer. Hence, each block has very rich multi-resolution features. 

\noindent{\emph{Learning one-third and reusing two-thirds.}} Fig.~\ref{recurrent} shows the building block that constructs our feature-pyramids. With coarser-resolution and fine-resolution features, we introduce a bottleneck block with a $1\times1$ conv-layer plus a $3\times3$ conv-layer to learn new features. The number of parameters is one-third compared with DSOD.

\noindent{\textbf{Concatenation with A Reduplicative Scheme.}}
Each concatenation operation merges feature maps of the same spatial resolution from the down-sampling pathway and the up-sampling pathway. The process is iterated until the coarsest resolution block is generated. For $320\times320$ input images, we use six resolutions of features for predicting objects. The finest resolution is $40\times40$ and the coarsest resolution is $2\times2$. To start the iteration, we simply choose two adjacent scales of the current resolution as the inputs of down-sampling and up-sampling pathways. 
We also apply an extra $160\times160$ resolution as the input of the down-sampling pathway for the finest resolution ($40\times40$) to improve the ability of detecting small objects. 

\begin{figure*}[t]
	\centering
	\includegraphics[width=0.9\textwidth]{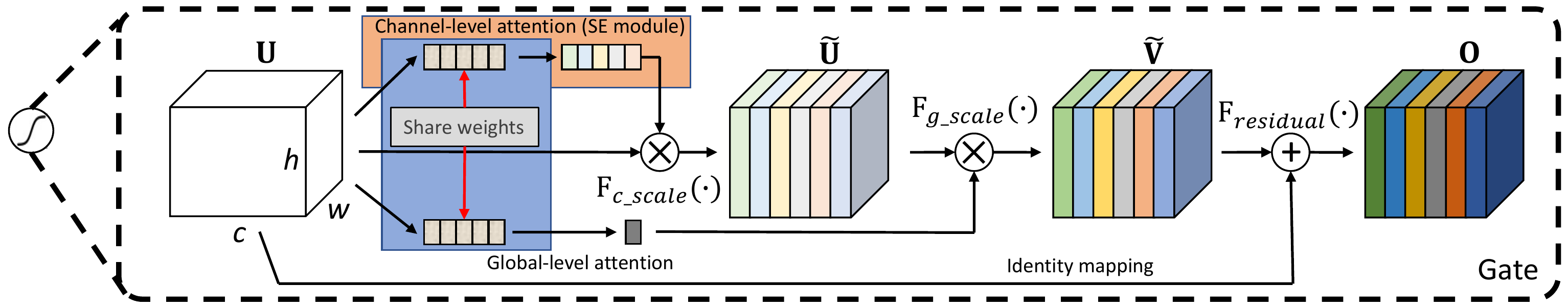}
	\vspace{0.1in}
	\caption {Illustration of the structure of a gate, including: (i) channel-level attention; (ii) global-level attention; and (iii) identity mapping. More details are given in \S\ref{sec:gate}.}
	\label{gate}
	\vspace{-0.1in}
\end{figure*}

\vspace{-1 ex}
\subsection{Gate-Controlled Adaptive Recalibration} \label{sec:gate}
\noindent{\textbf{Motivation.}}
Our goal is to ensure that the object detection network is able to adaptively select the meaningful scales for objects with different sizes, so that it can enhance the useful features in an appropriate resolution and suppress less useful ones. The  gate function $\mathbf{F}_{gate}$ can take any form, such as a fully-connected network or a convolutional network, but should be differentiable. In this paper, we propose to achieve this by using a two-level attention mechanism and an identity mapping before each prediction layer, partially inspired by Squeeze-and-Excitation~\cite{hu2017squeeze}.
A diagram of our gate structure is shown in Fig.~\ref{gate} and will be described in more detail in the following sections. We will verify the benefit of each component of our gate structure design in \S\ref{sec:ablation}.

\noindent{\textbf{Gate Definition.}}
A gate is a series of transformation $\mathbf{F}_{gate}$ that transforms the input feature maps $\mathbf{U}$ to outputs $\mathbf{O}$ ($\mathbf{U}\rightarrow\mathbf{O}$). Let $\mathbf{U}=[\mathbf{u}_1,\mathbf{u}_2,\dots,\mathbf{u}_c]$ denote the set of filter maps. Suppose $\mathbf{U}$, $\mathbf{O}$, $\widetilde{\mathbf{U}}$ and $\widetilde{\mathbf{V}}\in \mathbb{R}^{w \times h \times c}$ are all with width $w$, height $h$ and $c$ channels, where  $\widetilde{\mathbf{U}}$ and $\widetilde{\mathbf{V}}$ are intermediate states.
Denote $\mathbf{F}_{c\_scale}$ ($\mathbf{F}_{c}$ as abbreviation, similarly hereinafter): $\mathbf{U}\rightarrow\widetilde{\mathbf{U}}$; $\mathbf{F}_{g\_scale}$ ($\mathbf{F}_{g}$): $\widetilde{\mathbf{U}}\rightarrow\widetilde{\mathbf{V}}$ and $\mathbf{F}_{residual}$ ($\mathbf{F}_{r}$): $\widetilde{\mathbf{V}}\rightarrow{\mathbf{O}}$ and therefore a gate can be formulated as:
\begin{equation}
\mathbf{O} = {\mathbf{F}_{gate}}(\mathbf{U}) = {\mathbf{F}_{r}}({\mathbf{F}_{g}}({\mathbf{F}_{c}}(\mathbf{U})))
\end{equation}
\vspace{-3 ex}

\noindent{\textbf{Channel-level and Global-level Attention.}}
The aim of channel-level attention is to model relationships between channels and the global-level attention is to adaptively enhance or attenuate different scale supervision.
We apply Squeeze-and-Excitation block~\cite{hu2017squeeze} as our channel-level attention which consists of: (i) a {\em squeeze} stage $ {\mathbf{F}_{sq}}$ for global information embedding; and (ii) an {\em excitation} stage $ {\mathbf{F}_{ex}}$ for channel-level recalibration. Therefore we can formulate the channel-level outputs as:
\begin{equation}
\widetilde{\mathbf{U}} ={\mathbf{F}_{ex}}({\mathbf{F}_{sq}}(\mathbf{U}))
\end{equation}
The {\em squeeze} stage can be formulated as a global pooling operation on each channel:
\begin{equation}
{s_c} = {\mathbf{F}_{sq}}({\mathbf{u}_c}) = \frac{1}{{w \times h}}\sum\nolimits_{i = 1}^w {\sum\nolimits_{j = 1}^h {{u_c}} } (i,j)
\end{equation}
where $s_c$ is the $c$-th element of $\mathbf{s}$. $\mathbf{s}\in \mathbb{R}^c$ is a vector calculated by global-pooling filter $\mathbf{u}$. The {\em excitation} stage is two fully-connected layers plus a sigmoid activation:
\begin{equation}
\mathbf{e}=\mathbf{F}_{ex}(\mathbf{s})=\sigma(f_c(f_{\frac{c}{16}}(\mathbf{s})))
\end{equation}
where $\mathbf{e}\in \mathbb{R}^c$ is the output, $\sigma$ is the sigmoid function. $f_c$ and $f_{\frac{c}{16}}$ are the two fully-connected layers with output dimensions of $c$ and $\frac{c}{16}$, respectively. Then, we can calculate $\widetilde{\mathbf{U}}$ by:
\begin{equation}
\widetilde{\mathbf{U}}= {\mathbf{F}_{c}}(\mathbf{U}) = \mathbf{e}\otimes \mathbf{U}
\end{equation}
where $\otimes$ denotes channel-wise multiplication.
More details can be referred to the SENets~\cite{hu2017squeeze} paper.
Our global attention takes $\mathbf{s}$ (the output of {\em squeeze} stage) as input, and we modify the {\em excitation} stage by generating only one element. The new {\em excitation} stage ${\mathbf{\bar{F}}}_{ex}$ (for global attention) can be formulated as:
\begin{equation}
\mathbf{\bar e}=\mathbf{\bar F}_{ex}(\mathbf{s})=\sigma(f_\mathbf{{\textbf 1}}(f_{\frac{c}{16}}(\mathbf{s})))
\end{equation}
where $\mathbf{\bar e} \in \mathbb{R^{\textbf{1}}}$ is the global attention. The weight of $f_{\frac{c}{16}}$ is shared between ${\mathbf{F}_{ex}}$ and ${\mathbf{\bar{F}}}_{ex}$. Finally, $\widetilde{\mathbf{V}}$ is calculated by:
\begin{equation}
\widetilde{\mathbf{V}}= {\mathbf{F}_{g}}(\widetilde{\mathbf{U}}) = \mathbf{\bar e}\otimes \widetilde{\mathbf{U}}
\end{equation}

\begin{figure*}[t]
	\centering
	\includegraphics[width=0.8\textwidth]{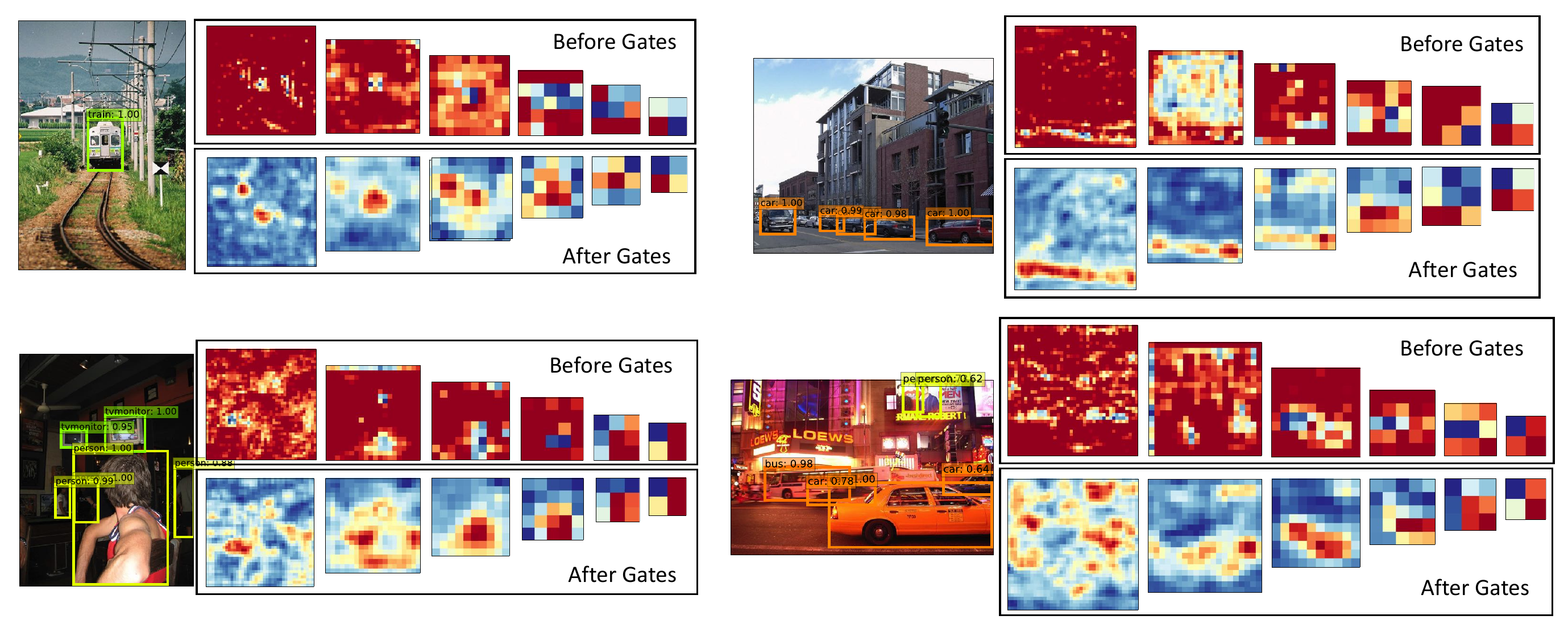}
	\vspace{0.05in}
	\caption{Visualization of feature maps \textbf{before} and \textbf{after} gates. In each block, left is the input image with  detection results. The right-top are the feature maps before gates and the right-bottom are the maps after gates.}
	\label{gate_visualization}
	\vspace{-0.1in}
\end{figure*}

\vspace{-3 ex}
\paragraph{Identity Mapping.} We use an element-wise addition operation~\cite{he2016deep} to obtain the final outputs:
\begin{equation}
\mathbf{O} = \mathbf{U} \oplus \widetilde{\mathbf{V}}
\end{equation}
where $\oplus$ denotes element-wise addition. 
Fig.~\ref{gate_visualization} shows several examples of feature map visualization before and after the gating operation.

\vspace{-1 ex}
\subsection{Feature Reuse for DSOD and SSD}
Our proposed method is a generic solution for building iterative feature pyramids and gates inside deep convolutional neural networks based detectors, thus it's very easy to apply to existing frameworks, such as SSD~\cite{liu2016ssd}, DSOD~\cite{Shen2017DSOD,8734700}, FPN~\cite{lin2016feature}, etc. In the following, we adopt our method on DSOD~\cite{Shen2017DSOD,8734700} and SSD~\cite{liu2016ssd} for general object detection, in order to demonstrate the effectiveness and advantages of our method. 

There are two steps to adapt {\em Gated Feature Reuse} for DSOD.
First, we apply {\em iterative feature reuse} to replace the dense connection in DSOD prediction layers. Following that, we add gates in each prediction layer to obtain GFR-DSOD. Other principles in DSOD are inherited in GFR-DSOD like {\em Stem}, {\em Dense Block}, etc. For SSD, similar operations are conducted to obtain GFR-SSD. Specifically, we  replace the extra layers in SSD with GFR structure and cascade gates in prediction layers.
Implementation details and empirical results are given in the next section.

\vspace{-1 ex}
\section{Experiments}

We conduct experiments on three widely used benchmarks: 20-category PASCAL VOC 2007, 2012~\cite{everingham2010pascal} and 80-category MS COCO detection datasets~\cite{lin2014microsoft}. Following the previous practice of learning detection from scratch~\cite{liu2016ssd,Shen2017DSOD,8734700}, 
we train using the union of VOC 2007 \texttt{trainval} and VOC 2012 \texttt{trainval} (``07+12'') and test on VOC 2007 test set. For VOC 2012,  we use VOC 2012 {\texttt {trainval}} and VOC 2007 {\texttt {trainval}} + {\texttt{test}} for training, and test on VOC 2012 {\texttt{test}} set.
For COCO with 80k images in training set, 40k in validation set and 20k in testing set ({\texttt {test-dev}}). 
In our study, all experiments are trained from scratch without ImageNet~\cite{deng2009imagenet} pre-trained models.
We adopt the backbone network proposed by DSOD~\cite{Shen2017DSOD} (GFR-DSOD) or VGGNet~\cite{simonyan2014very} (GFR-SSD) to ensure fair comparisons. 

\begin{table}[t]
	\centering
	\caption{Ablation Experiments of gate structure design on PASCAL VOC 2007.}
	\vspace{0.1in}
	\resizebox{0.5\textwidth}{!}{%
		\begin{tabular}{l|c}
			\hline
			Method  & mAP (\%) \\ \hline
			$+$ Channel-level attention & $+$0.4 (78.2)  \\ 
			$+$ Global-level attention  &  $+$0.2 (78.4) \\ 
			$+$ Identity mapping  & $+$0.2 (78.6)\\ \hline
		\end{tabular}
	}
	\label{ablation_gate}
		\vspace{-0.1in}
\end{table}	

\renewcommand{\arraystretch}{1.03}
\setlength{\tabcolsep}{1.0em}
\begin{table}[t]
	\centering
	\caption{Ablation Experiments on PASCAL VOC 2007. ``IFP'' denotes our iterative feature pyramids. We add additional aspect ratio 1.6 for default boxes at every prediction layer. While for DSOD, we found there was no improvement when adding more aspect ratios.}
	\vspace{0.1in}
	\resizebox{0.6\textwidth}{!}{%
		\begin{tabular}{l|c}
			\hline
			Method & mAP (\%) / \#params \\ \hline
			DSOD300~\cite{Shen2017DSOD}	&  77.7 (14.8M)  \\ 
			GFR-DSOD300&  \textbf{78.9 (14.1M)}   \\ \hline\hline
			DSOD320~\cite{Shen2017DSOD} & {77.8} \\ 
			DSOD320 (using FR only)&  78.6   \\ 
			DSOD320 (using gates only)&  78.6   \\ \hline
			GFR-DSOD320& \textbf{79.2}    \\ \hline
		\end{tabular}
	}
	\label{ablation_dsod}
	
		\centering
	\caption{Ablation Experiments of SSD300 from scratch on PASCAL VOC 2007. } 
	\vspace{0.12in}
	\resizebox{0.65\textwidth}{!}{%
		\begin{tabular}{l|c|c}
			\hline
			Method & \# params & mAP (\%) \\ \hline
			SSD300~\cite{Shen2017DSOD}	& 26.3M &69.6  \\ 
			SSD300 (dense pred.)~\cite{Shen2017DSOD}	& 26.0M & 70.4  \\ 
			GFR-SSD300 & \textbf{24.9M} & \textbf{73.2}\\ \hline
		\end{tabular}
	}
	\label{ablation_ssd}
	\vspace{-0.14in}
	
\end{table}

\begin{table*}[t]
	\centering
	\caption{\textbf{Comparisons of two-stage detectors on MS COCO 2015 \texttt{test-dev} set.}}
	\label{COCO}
	\vspace{0.1in}
	\resizebox{0.99\textwidth}{!}{%
		\begin{tabular}{l|c|c|c|ccc}
			\hline
			\multirow{2}{*}{Method}          & \multirow{2}{*}{\textbf{network}} & \multirow{2}{*}{\textbf{pre-train}} & \multirow{2}{*}{\textbf{\# param}} & \multicolumn{3}{c}{\textbf{COCO (Avg. Precision, IoU:)}}  \\
			&                       &              &            & \textbf{0.5:0.95}        & \textbf{0.5}        & \textbf{0.75}          \\ \hline
			\multicolumn{7}{l}{\textbf{One-Stage Detectors:}}        \\ \hline
			\textbf{SSD300}~\cite{liu2016ssd}   &  VGGNet  & \Checkmark & 34.3M & 23.2      &  41.2    &  23.4      \\ 
			\textbf{SSD320*}~\cite{liu2016ssd}   &  VGGNet  & \Checkmark & 34.3M & 25.1           & 43.1       & 25.8       \\ 
			\textbf{DSSD321}~\cite{fu2017dssd}   &  ResNet-101  & \Checkmark & 256M & 28.0           & 46.1       & 29.2       \\ 
			\textbf{DSOD320}~\cite{Shen2017DSOD}   &   DSOD   & \ding{55}  & 21.9M & 29.4            & 47.5       & 30.7       \\ 
			\textbf{Ours (GFR-DSOD320)}   &  DSOD & \ding{55} & \textbf{21.2M} & \textbf{30.0}  & 47.9   & \textbf{31.8}  \\ \hline \hline
			\multicolumn{7}{l}{\textbf{Two-Stage Detectors:}}        \\ \hline
			\textbf{FRCNN320/540}~\cite{ren2015faster}   &  ResNet-101  & \Checkmark & 62.6M &     23.3    &    42.8   &  23.1   \\ 
			\textbf{R-FCN320/540+OHEM}~\cite{dai2016r}   &  ResNet-101  & \Checkmark & 54.4M &    25.8     &   45.8    &   26.4  \\ 
			\textbf{Deformable FRCNN320/540}~\cite{dai2017deformable,ren2015faster} &  ResNet-101  & \Checkmark & 64.8M &    28.3     &   46.5    &   30.2  \\ 
			\textbf{Deformable R-FCN320/540}~\cite{dai2017deformable} &  ResNet-101  & \Checkmark & 63.6M &    29.5     &  47.6     &  30.7   \\ 
			\textbf{FPN320/540}~\cite{lin2016feature}   &  ResNet-101  & \Checkmark & 121.2M &     29.7     &   \textbf{48.1}    &    31.0   \\ 
			\hline 
		\end{tabular}
	}
	\vspace{-0.1in}
\end{table*}

\noindent{\textbf{Implementation details.}}
We adopt SGD for training our models on 8 GPUs. Following~\cite{Shen2017DSOD,liu2016ssd}, we use a weight decay of 0.0005 and a momentum of 0.9. All conv-layers are initialized with the ``xavier'' method~\cite{glorot2010understanding}.
For other settings, we followed the same implementation as in the original DSOD~\cite{Shen2017DSOD} and SSD~\cite{liu2016ssd} papers.

\subsection{Ablation Experiments on PASCAL VOC 2007} \label{sec:ablation}

In this section, we investigate the effectiveness of each component of our GFR-DSOD framework. We design several controlled experiments on PASCAL VOC 2007 for the ablation study, including: (i) iterative feature pyramids; (ii) gates; and (iii) two level attention and identity mapping. In these experiments, we train on the union of VOC 2007 \texttt{trainval} and 2012 \texttt{trainval} (``07+12''), test on the VOC 2007 \texttt{test} set.

\noindent{\textbf{Effectiveness of Channel attention, Global attention and Identity mapping.}}
 Tab.~\ref{ablation_gate} shows the ablation results of gate structure design. After adopting channel attention, global attention and identity mapping, we obtain gains of 0.4\%, 0.2\% and 0.2\%, respectively.

\noindent{\textbf{Effectiveness of Iterative Feature Reuse.}}
Tab.~\ref{ablation_dsod} (row 4) shows the results of our feature pyramids without the gates. The result (78.6\%) is on par with GFR-DSOD320 (row 6) and achieves 0.8\% improvement comparing with baseline (77.8\%). It indicates that our feature reuse structure contribute a lot on boosting the final detection performance. 

\noindent{\textbf{Effectiveness of Gates.}}
Tab.~\ref{ablation_dsod} (row 5) shows the results of adding gates without the iterative feature pyramids. The result (78.6\%) also outperforms the baseline result by 0.8\% mAP.

We also show GFR-SSD results in Tab.~\ref{ablation_ssd}. We can see that our GFR structure helps the original SSD to improve the performance by a large margin. We conjecture the reasons are two-fold: First, the baseline (69.6\%) is still at a very low level. So there is a very large room for performance improvement. Second, VGGNet backbone is a plain network, our GFR structure (skip connection from very low-level to high-level layers) helps this kind of structure greatly. After using more channels and batch norm~\cite{ioffe2015batch} in each GFR pyramid, our result is further improved to \textbf{75.8\% ($\uparrow 6.2\%$ mAP)} with only \textbf{25.4M} model parameters.

\vspace{-1 ex}
\subsection{Results on PASCAL VOC 2007\&2012} \label{sec:voc07}
Tab.~\ref{ablation_dsod} shows our results on VOC2007 \texttt{test} set. Our GFR-DSOD achieves 79.2\%, which is better than baseline method DSOD (77.8\%).
Fig.{\color{red}{7}} shows some qualitative detection examples on VOC 2007  \texttt{test} set with DSOD and our GFR-DSOD models. Our method achieves better results on both small objects and dense scenes

\noindent{\textbf{Convergence Speed Comparison.}}
In GFR-DSOD, we observe that models always obtain the best accuracy with 62k iterations. In DSOD, however, the models need around 100k iterations to achieve final convergence, with the same batch size. Thus, GFR-DSOD has relative 38\% faster convergence speed than DSOD. Fig. {\color{red}{6}} shows the comparison of training and testing accuracy with DSOD method. For the inference time, With $300\times300$ input, our full GFR-DSOD can run an image at 17.5 {\em fps} on a single Titan X GPU with batch size 1. The speed is similar to DSOD300 with the dense prediction structure. When enlarging the input size to $320\times320$, the speed decrease to 16.7 {\em fps} and 16.3 {\em fps} (with more default boxes). As comparisons, SSD321 runs at 11.2 {\em fps} and DSSD321 runs at 9.5 {\em fps} with ResNet-101~\cite{he2016deep} backbone network. Our method is much faster than these two competitors.

\vspace{2 ex}
\begin{minipage}[b]{0.35\linewidth}
	\includegraphics[height=1\textwidth]{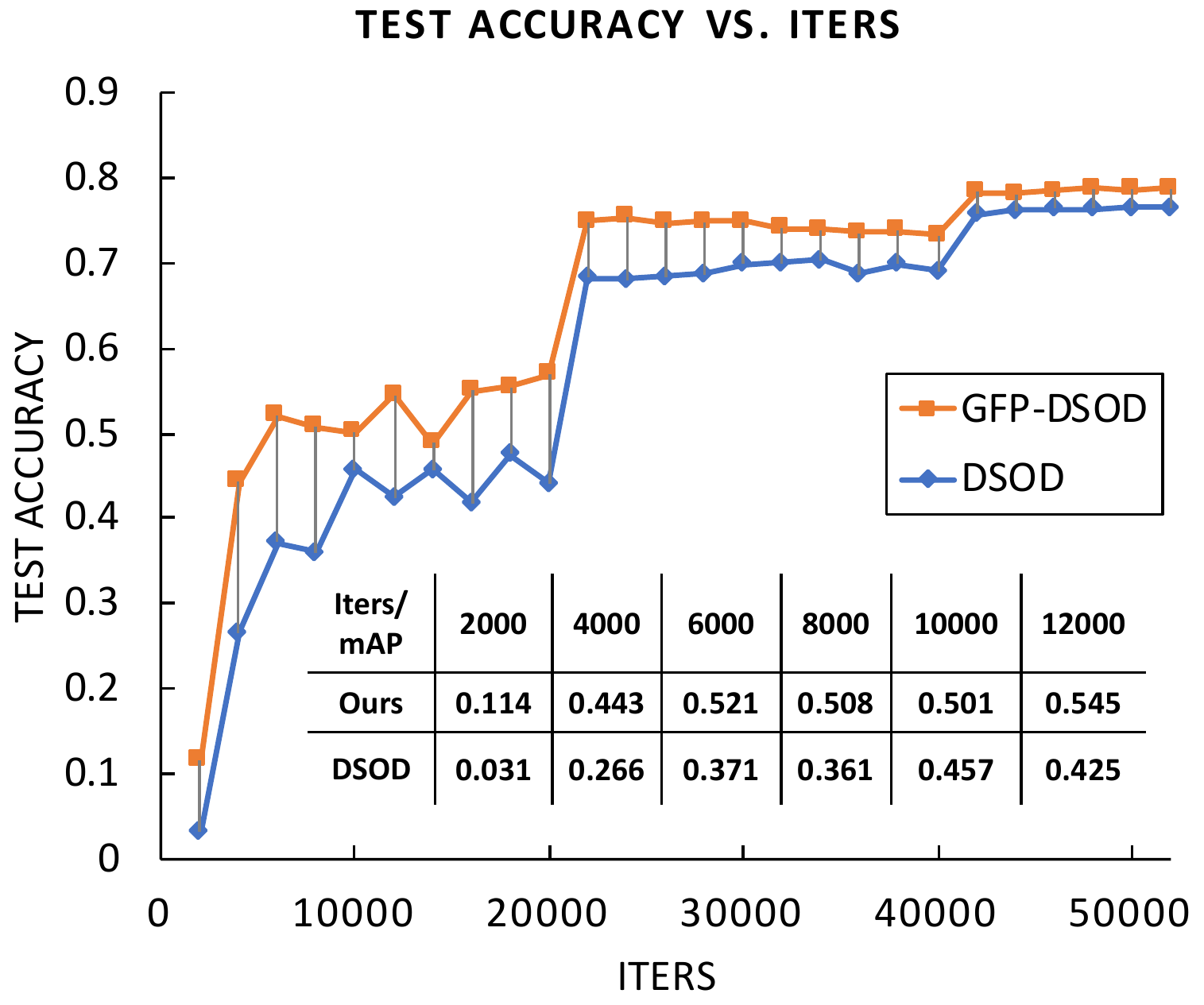}
\end{minipage}
\hfill
\begin{minipage}[b]{0.5\linewidth}
	{Figure 6: Comparison of DSOD~\cite{Shen2017DSOD} (blue diamonds) and GFR-DSOD (orange squares) on VOC 2007 \texttt{test} set. We show results (mAP) of GFR-DSOD and DSOD at six different iterations. GFR-DSOD obtains much higher accuracy with same training iterations and also achieves faster convergence than the baseline DSOD. Details are given in \S\ref{sec:voc07}.}
	\label{acc_c}
	\vspace{0.3in}
\end{minipage}

\begin{minipage}[b]{0.38\linewidth}
	\includegraphics[height=1.2\textwidth]{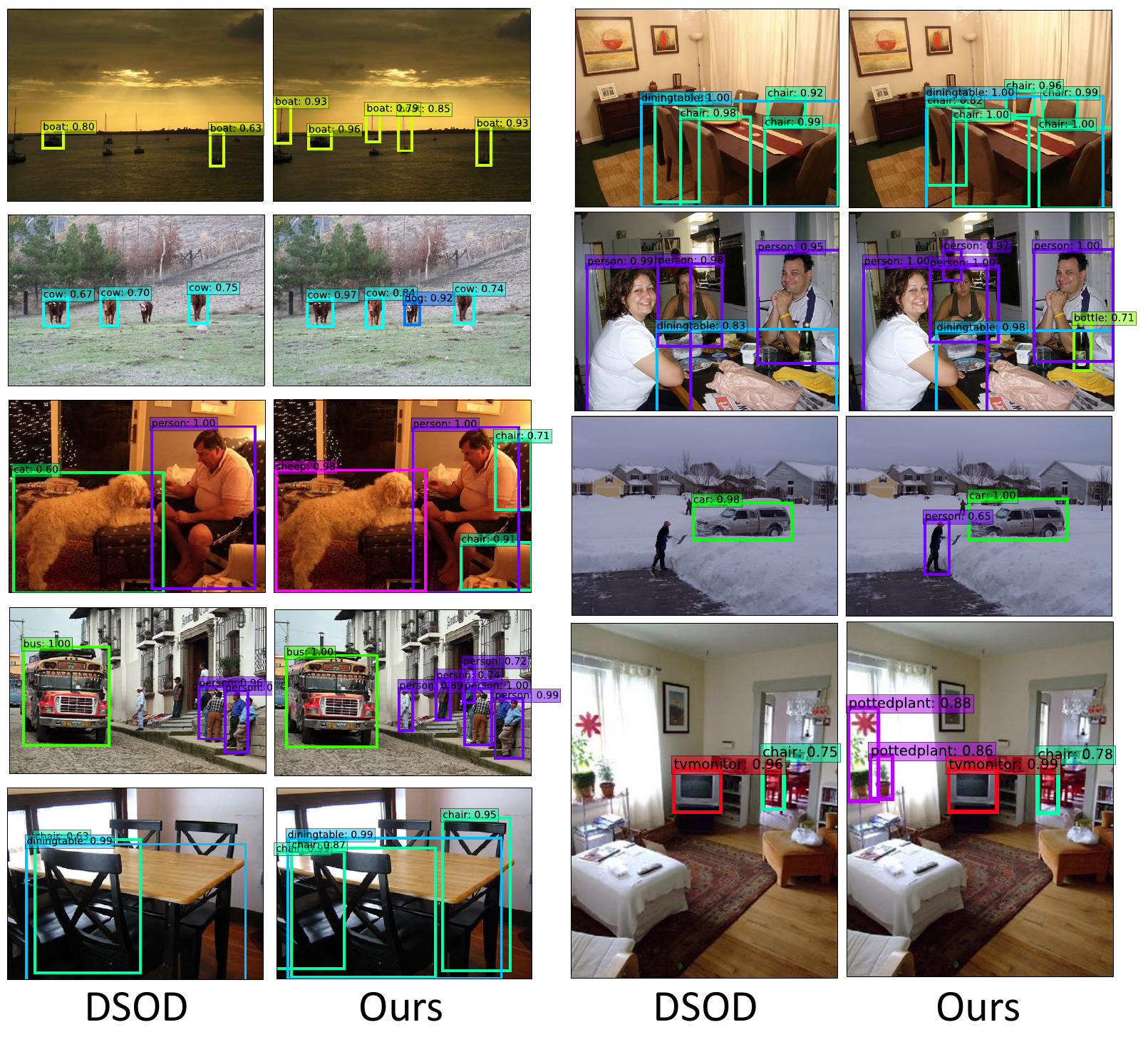}
\end{minipage}
\hfill
\begin{minipage}[b]{0.42\linewidth}
	{Figure 7: Detection examples on VOC 2007 \texttt{test} set with DSOD / GFR-DSOD models. For each pair, the left is the result of DSOD and right is the result of GFR-DSOD. We show detections with scores higher than 0.6. Each color corresponds to an object category in that image. Our method achieves better performance on both small objects and dense scenes.}
	\label{examples}
	\vspace{0.4in}
\end{minipage}

We also perform our experiments on VOC 2012 \texttt{test} set with two different training subsets of related experimental results: one with models only trained from scratch with VOC 2012 provided data (VOC 2012 \texttt{Comp3} Challenge), and another one with additional VOC 2007 data for training. 
On PASCAL VOC 2012 \texttt{Comp3} Challenge, our result (72.5\%) outperforms the previous state-of-the-art DSOD (70.8\%)~\cite{Shen2017DSOD} by 1.7\% mAP.  After adding VOC 2007 as training data, our GFR-DSOD320 (77.5\% mAP) is consistently better than baseline DSOD320 (76.4\%) and some other state-of-the-art methods using VGGNet or ResNet-101 pre-trained models like SSD321 (75.4\%) and DSSD321 (76.3\%).

\vspace{-1 ex}
\subsection{Results on MS COCO}
Finally we evaluate our GFR-DSOD on the MS COCO dataset~\cite{lin2014microsoft}. 
The batch size is set to 128.
Results are summarized in Tab.~\ref{COCO}. 
We first compare our method with state-of-the-art one-stage detectors.
We observe that GFR-DSOD can achieve higher performance than the baseline method DSOD (30.0\% {\em vs.} 29.4\%) with fewer parameters (21.2M {\em vs.} 21.9M).
We then compare our method with state-of-the-art two-stage detectors. It can be observed that our model is much more compact than that of the two-stage detectors using ResNet-101 as backbone. For instance, our results is comparable with FPN320/540~\cite{lin2016feature} (30.0\% {\em vs.} 29.7\%), but the parameters of our model is only 1/6 of FPN.

\section{Conclusions}

We have presented \textit{gated feature reuse}, a novel structure deign that is more fit for learning detection from scratch, 
and improves performance in model parameters, convergence speed and accuracy.
Extensive experiments on PASCAL VOC 2007, 2012 and MS COCO demonstrate the effectiveness of our proposed method. Since learning from scratch uses limited training data, our future work will focus on adopting GAN-based image generation and data augmentation method like MUNIT~\cite{huang2018multimodal}, INIT~\cite{shen2019towards}, etc. to enlarge the diversity of training objects, in order to obtain better detection performance when learning form scratch.

\bibliography{GFR}
\end{document}